\documentclass{article}

\PassOptionsToPackage{numbers, compress}{natbib}
\usepackage[final]{nips_2017}

\usepackage[utf8]{inputenc} % allow utf-8 input
\usepackage[T1]{fontenc}    % use 8-bit T1 fonts
\usepackage{hyperref}       % hyperlinks
\usepackage{url}            % simple URL typesetting
\usepackage{booktabs}       % professional-quality tables
\usepackage{amsfonts}       % blackboard math symbols
\usepackage{nicefrac}       % compact symbols for 1/2, etc.
\usepackage{microtype}      % microtypography

\usepackage{xr}
\usepackage{natbib}
\usepackage{graphicx}

\usepackage{amsthm}
\usepackage{amsmath}
\usepackage{enumitem}
\usepackage{hyperref}    
\usepackage{algorithm}
\usepackage{algorithmic}
\newtheorem{theorem}{Theorem}
\newtheorem{example}{Example}
\newtheorem{property}{Property}
\newtheorem{proposition}{Proposition}
\newtheorem{definition}{Definition}

\usepackage{tablefootnote}
\usepackage{multicol}
\usepackage{subfig}
\usepackage{pdfpages}

\makeatletter
\newcommand*{\indep}{%
  \mathbin{%
    \mathpalette{\@indep}{}%
  }%
}
\newcommand*{\nindep}{%
  \mathbin{%                   % The final symbol is a binary math operator
    \mathpalette{\@indep}{\not}% \mathpalette helps for the adaptation
  }%
}
\newcommand*{\@indep}[2]{%
  \sbox0{$#1\perp\m@th$}%        box 0 contains \perp symbol
  \sbox2{$#1=$}%                 box 2 for the height of =
  \sbox4{$#1\vcenter{}$}%        box 4 for the height of the math axis
  \rlap{\copy0}%                 first \perp
  \dimen@=\dimexpr\ht2-\ht4-.2pt\relax
  \kern\dimen@
  {#2}%
  \kern\dimen@
  \copy0 %                       second \perp
} 
\makeatother

\title{Random Subspace with Trees for Feature Selection Under Memory Constraints}

\author{
  Antonio Sutera\\
  University of Liège, Belgium\\
  \texttt{sutera.antonio@gmail.com} \\
  \And
  Célia Châtel\\% Coauthor \\
  Aix-Marseille University, France%% Affiliation \\
  \And
  Gilles Louppe\\%% Coauthor \\
  New York University, USA\\%% Affiliation \\
  \And
  Louis Wehenkel\\%% Coauthor \\
  University of Liège, Belgium\\%% Affiliation \\
  \And
  Pierre Geurts\\%% Coauthor \\
  University of Liège, Belgium%% Affiliation \\
}

\newcommand{\beforesection}[0]{\vspace{-0.4em}}
\newcommand{\aftersection}[0]{\vspace{-0.4em}}

\newcommand{\beforesubsection}[0]{\vspace{-0.4em}}
\newcommand{\aftersubsection}[0]{\vspace{-0.5em}}

\newcommand{\beforeparagraph}[0]{\vspace{-0.7em}}

\newcommand{\afterformula}[0]{\vspace{-0.4em}}
\begin{document}

\maketitle

\begin{abstract} \vspace{-0em}
Dealing with datasets of very high dimension is a major challenge in machine
learning. In this paper, we consider the problem of feature selection in
applications where the memory is not large enough to contain all features. In
this setting, we propose a novel tree-based feature selection approach that
builds a sequence of randomized trees on small subsamples of variables mixing
both variables already identified as relevant by previous models and variables
randomly selected among the other variables. As our main contribution, we
provide an in-depth theoretical analysis of this method in infinite sample
setting. In particular, we study its soundness with respect to common
definitions of feature relevance and its convergence speed under various
variable dependance scenarios. We also provide some preliminary empirical
results highlighting the potential of the approach.
\end{abstract}

\vspace{-0.3em}\section{Motivation}\aftersection

We consider supervised learning and more specifically feature selection in
applications where the memory is not large enough to contain all data. Such
memory constraints can be due either to the large volume of available training
data or to physical limits of the system on which training is performed (eg.,
mobile devices). A straightforward, but often efficient, way to handle such
memory constraint is to build and average an ensemble of models, each trained
on only a random subset of samples and/or features that can fit into
memory. Such simple ensemble approaches have the advantage to be applicable to
any batch learning algorithm, considered as a black-box, and they have been
shown empirically to be very effective in terms of predictive performance, in
particular when combined with trees, and even when samples and/or features are
selected uniformly at random \citep[see,
  eg.,][]{chawla2004,louppe2012ensembles}.  In particular, and independently of
any considerations about memory constraints, feature subsampling has been shown
in several works to be a very effective way to introduce randomization when
building ensembles of models \citep{ho1998random,kuncheva2010random}. The idea
of feature subsampling has also been investigated in the context of feature
selection, where several authors have proposed to repeatedly apply a
multivariate feature selection technique on random subsets of features and then
to aggregate the results obtained on these subsets \citep[see,
  eg.,][]{draminski2008monte,lai2006random,konukoglu2014approximate,nguyen2015new,draminski2016discovering}.

% We propose in this work an improved version of the Random Subspace method (SRS). However, the main contribution of this paper lies in the theoretical analysis of this method in asymptotic setting that shows (1) that our algorithm is able to find all (strongly) relevant features in most circumstances, (2) that it can improve convergence speed with respect to Random Subspace (RS) by several orders of magnitude in some scenarios and (3) that these scenarios are relevant for a large class of (PC) distributions (theorem 3). 

% As an important additional contribution, we believe that our analysis also sheds some new light on both the popular random subspace and random forests methods that are special cases of the SRS algorithm. For example, we highlight the conditions under which these methods are able to identify all (strongly) relevant features.

% We believe that our theoretical analysis will shed some new light on these latter studies.

In this work, focusing on feature subsampling, we adopt a simplistic setting
where we assume that only q input features (among $p$ in total, with typically
$q \ll p$) can fit into memory. In this setting, we study ensembles of
randomized decision trees trained each on a random subset of $q$ features. In
particular, we are interested in the properties of variable importance scores
derived from these models and their exploitation to perform feature selection.
In contrast to a purely uniform sampling of the features, we propose in
Section~\ref{sec:srs} a modified sequential random subspace (SRS) approach that
biases the random selection of the features at each iteration towards features
already found relevant by previous models. As our main contribution, we perform
in Section~\ref{sec:analysis} an in-depth theoretical analysis of this method
in infinite sample size condition. In particular, we show that (1) this
algorithm provides some interesting asymptotic guarantees to find all
(strongly) relevant variables, (2) that accumulating previously found variables
can reduce the number of trees needed to find relevant variables by several
orders of magnitudes with respect to the standard random subspace method in
some scenarios, and (3) that these scenarios are relevant for a large class of
(PC) distributions. As an important additional contribution, our analysis also
sheds some new light on both the popular random subspace and random forests
methods that are special cases of the SRS algorithm. Finally,
Section~\ref{sec:empirical} presents some preliminary empirical results with
the approach on several artificial and real datasets.

\beforesection
\section{Feature selection and tree-based methods}
\label{sec:background}
\aftersection
% Background feature selection

% Background classique arbres et importances
% In this work, we consider decision tree as standard base estimators however...

This section gives the necessary background about feature selection
and random forests.

\beforesubsection
\subsection{Feature relevance and feature selection}\label{sec:relevance}\aftersubsection

Let us denote by $V$ the set of inputs variables, with $|V|=p$, and by
$Y$ the output. Feature selection is concerned about the
identification in $V$ of the (most) relevant variables. A common
definition of relevance is as follows \citep{kohavi1997wrappers}:
\begin{definition}
A variable $X\in V$ is {\bf relevant} iff there exists a subset $B\subset
V$ such that $X\nindep Y|B$. A variable is {\bf irrelevant} if it is
not relevant.
\end{definition} \afterformula
Relevant variables can be further divided into two categories
\citep{kohavi1997wrappers}:
\begin{definition}
A variable $X$ is {\bf strongly relevant} iff $Y\nindep
X|V\setminus\{X\}$. A variable $X$ is {\bf weakly relevant} if it is
relevant but not strongly relevant.
\end{definition} \afterformula
Strongly relevant variables are thus those variables that convey
information about the output that no other variable (or combination of
variables) in $V$ conveys.

The problem of feature selection usually can take two flavors
\citep{nilsson2007consistent}:\vspace{-0.7em}
\begin{itemize}[leftmargin=*]
\item \textbf{All-relevant problem}: finding all relevant features.
\item \textbf{Minimal optimal problem}: finding a subset $M\subseteq
  V$ such that $Y\indep V\setminus M|M$ and such that no proper
  subset of $M$ satisfies this property. A subset $M$ solution to the
  minimal optimal problem is called a {\bf Markov boundary} (of $Y$
  with respect to $V$).
\end{itemize}\vspace{-3mm}
A Markov boundary always contains all strongly relevant variables and
potentially some weakly relevant ones. In general, the minimal optimal
problem does not have a unique solution.  For strictly positive
distributions\footnote{Following \citep{nilsson2007consistent}, we
  will define a strictly positive distribution $P$ over $V\cup\{Y\}$
  as a distribution such that $P(V=v)>0$ for all possible values $v$
  of the variables in $V$.}  however, the Markov boundary $M$ of $Y$
is unique and a feature $X$ belongs to $M$ iff $X$ is strongly
relevant \citep{nilsson2007consistent}. In this case, the solution to
the minimal optimal problem is thus the set of all strongly relevant
variables.

In what follows, we will need to qualify relevant variables according
to their degree:
\begin{definition}
The {\bf degree} of a relevant variable $X$, denoted $deg(X)$, is
defined as the minimal size of a subset $B\subseteq V$ such that
$Y\nindep X|B$.
\end{definition} \afterformula
Relevant variables $X$ of degree 0, i.e. such that $Y\nindep X$
unconditionally, will be called {\bf marginally relevant}.

We will say that a subset $B$ such that $Y\nindep
X|B$ is {\bf minimal} if there is no proper subset $B'\subseteq B$
such that $Y\nindep X|B'$. The following two propositions give a
characterization of these minimal subsets.
\begin{proposition} \label{prop:only-relevant}
A minimal subset $B$ such that $Y\nindep X|B$ for a relevant
  variable $X$ contains only relevant variables. \quad (Proof in Appendix~\ref{app:only-relevant})
\end{proposition}
% \begin{proof}
% See Appendix~\ref{app:only-relevant}
% \end{proof}\vspace{-0.5em}

\begin{proposition} \label{prop:only-degree}
Let $B$ denote a minimal subset such that $Y\nindep X|B$ for a
relevant variable $X$. For all $X'\in B$, $deg(X')\leq |B|$. \quad (Proof in Appendix~\ref{app:only-degree})
\end{proposition}\afterformula
% \begin{proof}
%     See Appendix~\ref{app:only-degree}
% \end{proof}\vspace{-1em}
These two propositions show that a minimal conditioning $B$ that makes a
variable dependent on the output is composed of only relevant variables whose
degrees are all smaller or equal to the size of $B$. We will provide in Section
4.2 a more stringent characterization of variables in minimum conditionings in
the case of a specific class of distributions.

\beforesubsection
\subsection{Tree-based methods and variable importances}
\aftersubsection

%% We review here single decision trees and random forests algorithms. We
%% then present variable importance measures derived from these models.

%\beforeparagraph\paragraph{Decision trees.}
A {\bf decision tree} \citep{breiman1984cart} represents an input-ouput model with a
tree structure, where each interior node is labeled with a test based
on some input variable and each leaf node is labeled with a value of
the output. The tree is typically grown using a recursive procedure
which identifies at each node $t$ the split $s$ that maximizes the
mean decrease of some node impurity measure (e.g., Shannon entropy in classification and variance in regression). %% , as defined by (in the case of a
%% binary split):
%% $$\Delta i (s,t) = i(t) - p_L i(t_L)-p_R i(t_R),$$ where $i$ is some impurity
%% measure that depends on the nature of the output 
%%(e.g., Shannon entropy in classification and variance in regression),
%% $t$ is the node to split, $t_L$ and $t_R$ are its left and right
%% successors after the split, and $p_L$ and $p_R$ are the proportion of
%% examples from $t$ that fall into these successors. When trees are
%% fully developed, the split of a node is stopped only when either the
%% output or all inputs are locally constant in the subsample reaching
%% that node.

%% \beforeparagraph\paragraph{Ensemble of randomized trees} 
%% Typically, decision trees suffer from a high variance, which is
%% reduced by building an ensemble of randomized trees and aggregating
%% their predictions. More precisely, some randomization is introduced in
%% the tree learning procedure in order to produce a set of different
%% trees from a single learning set. Several techniques have been
%% proposed in the literature to introduce this randomization. For
%% example, bagging \citep{breiman1996bagging} builds each tree with the
%% classical algorithm from a bootstrap sample from the original learning
%% sample. \cite{ho1998random}'s random subspace method grows each tree
%% from a subset of the features of size $K\leq p$ randomly drawn from
%% $V$. \citet{breiman2001random}'s Random Forests algorithm combines
%% bagging with a {\it local} random selection of $K(\leq p)$ variables
%% at each node from which to identify the best split.

Typically, decision trees suffer from a high variance that can be
very efficiently reduced by building instead an {\bf ensemble of randomized trees}
and aggregating their predictions.  Several techniques have been
proposed in the literature to grow randomized trees. For
example, bagging \citep{breiman1996bagging} builds each tree with the
classical algorithm from a bootstrap sample from the original learning
sample. \citet{ho1998random}'s random subspace method grows each tree
from a subset of the features of size $K\leq p$ randomly drawn from
$V$. \citet{breiman2001random}'s Random Forests algorithm combines
bagging with a {\it local} random selection of $K(\leq p)$ variables
at each node from which to identify the best split.

%\beforeparagraph\paragraph{Variable importances.}
Given an ensemble of trees, several methods have been proposed to evaluate the
{\bf importance} of the variables for predicting the output
\citep{breiman1984cart,breiman2001random}. We will focus here on one particular measure
called the mean decrease impurity (MDI) importance for which some theoretical
characterization has been proposed in \citep{louppe13-nips}. This measure
adds up the weighted impurity decreases %$p_t \Delta i(s_t,t)$
over all nodes $t$ in a tree $T$ where the variable $X$ to score is used to split and
then averaging this quantity over all trees in the ensemble, i.e.:
\begin{equation}\label{mdi-imp}
  Imp(X) = \frac{1}{N_T} \sum_T \sum_{t\in T:v(s_t)=X} p(t) \Delta
i(s_t,t),\mbox{ with } \Delta i (s_t,t) = i(t) - p(t_L) i(t_L)-p(t_R) i(t_R)
\end{equation}
where $i$ is the impurity measure, $p(t)$ is the proportion of samples
reaching node $t$, $v(s_t)$ is the variable used in the split $s_t$ at
node $t$, and $t_L$ and $t_R$ are the left and right successors of $t$ after the
split.

\citet{louppe13-nips} derived several interesting properties of this measure
under the assumption that all variables are discrete and that splits on these
variables are multi-way (i.e., each potential value of the splitting variable
is associated with one successor of the node to split). In particular, they
obtained the following result in asymptotic sample and ensemble size
conditions:
\begin{theorem}\label{th:louppe}
  $X\in V$ is irrelevant to $Y$ with respect to $V$ if and only if its infinite
  sample importance as computed with an infinite ensemble of fully developed
  totally randomized trees built on $V$ for $Y$ is 0 (Theorem 3 in \citet{louppe13-nips}).
\end{theorem} \afterformula
Totally randomized trees are trees obtained by setting Random Forests
randomization parameter $K$ to 1.  This result shows that MDI
importance derived from trees grown with $K=1$ is asymptotically
consistent with the definition of variable relevance given in the
previous section. In Section \ref{sec:soundness}, we will actually
extend this result to values of $K$ greater than 1.

\beforesection
\section{Sequential random subspace}\label{sec:srs}
\aftersection
\begin{algorithm}[t]%[h]
   
  \footnotesize
  \caption{Sequential Random Subspace algorithm}
  \label{algo:SRS}
  %\noindent~\hrulefill\\
  {\bf Inputs:}\\
  \underline{Data}: $Y$ the output and $V$, the set of all input variables (of size $p$).\\
  \underline{Algorithm}: $q$, the subspace size, and $T$ the number of iterations, $\alpha\in[0,1]$, the percentage of memory devoted to previously found features.\\
  \underline{Tree}: $K$, the tree randomization parameter\\
  {\bf Output:} An ensemble of $T$ trees and a subset $F$ of features\\
  {\bf Algorithm:}\vspace{-2mm}
  \begin{enumerate}[leftmargin=*]
    \setlength\itemsep{0em}
    \item $F=\emptyset$
    \item Repeat $T$ times:
    \begin{enumerate}[leftmargin=*,topsep=0pt]
      %    \item Let $R$ be a subset of $\min\{\lfloor \alpha q\rfloor, |F|\}$ features randomly picked in $F$ without replacement.
      %    \item Let $C$ be a subset of $q-|R|$ features randomly selected in $V\setminus R$.
      \item Let $Q=R\cup C$, with $R$ a subset of $\min\{\lfloor \alpha
      q\rfloor, |F|\}$ features randomly picked in $F$ without
      replacement and $C$ a subset of $q-|R|$ features randomly
      selected in $V\setminus R$.
      \item Build a decision tree $\cal T$ from $Q$ using randomization parameter $K$.
      \item Add to $F$ all features from $Q$ that get an importance
      %(significantly$^*$) 
      greater than zero in $\cal T$.
    \end{enumerate}
  \end{enumerate}
  \vspace{-0mm}
  %\textit{$^*$Significance can be tested by random permutations.}\\
  \vspace{-2mm}
  %\noindent~\hrulefill
  \vspace{-0mm}
\end{algorithm}

In this paper, we consider a simplistic memory-constrained setting where it is
assumed that only $q$ input features can fit into memory at once, with
typically $q$ small with respect to $p$. Under this hypothesis,
Algorithm~\ref{algo:SRS} describes the proposed sequential random subspace (SRS)
algorithm to build an ensemble of randomized trees, which generalizes the
Random Subspace (RS) method \cite{ho1998random}. The idea of this method is to
bias the random selection of the features at each iteration towards features
that have already been found relevant by the previous trees. A parameter
$\alpha$ is introduced that controls the degree of accumulation of previously
identified features. When $\alpha=0$, SRS reduces to the standard RS
method. When $\alpha=1$, all previously found features are kept while when
$\alpha<1$, some room in memory is left for randomly picked features, which
ensures some permanent exploration of the feature space. Further randomization
is introduced in the tree building step through the parameter $K\in [1,q]$,
ie. the number of variables sampled at each tree node for splitting. Variable
importance is assumed to be the MDI importance. This algorithm returns both an
ensemble of trees and a subset $F$ of variables, those that get an importance
(significantly) greater than 0 in at least one tree of the ensemble. Importance
scores for the variables can furthermore be derived from the final ensemble
using (\ref{mdi-imp}). In what follows, we will denote by $F^{K,\alpha}_{q,T}$
and $Imp^{K,\alpha}_{q,T}(X)$ resp. the set of features and the importance of
feature $X$ obtained from an ensemble grown with SRS with parameters
$K$, $\alpha$, $q$ and $T$.

The modification of the RS algorithm is actually motivated by Propositions
\ref{prop:only-relevant} and \ref{prop:only-degree}, stating that the relevance
of high degree features can be determined only when they are analysed jointly
with other relevant features of equal or lower degree. From this result, one
can thus expect that accumulating previously found features will fasten the
discovery of higher degree features on which they depend through some snowball
effect. In the next section, we provide a theoretical asymptotic analysis of
the SRS method that confirms and quantifies this effect.

Note that the SRS method can also be motivated from the perspective of
accuracy.  When $q\ll p$ and the number of relevant features $r$ is also much
smaller than the total number of features $p$ ($r\ll p$), many trees with
standard RS are grown from subsets of features that contain only very few, if
any, relevant features and are thus expected not to be better than random
guessing \cite{kuncheva2010random}. In such setting, RS ensembles are thus
expected not to be very accurate.
\begin{example} With $p=10000$, $r=10$ and $q=50$, the proportion of trees in a RS ensemble grown from only irrelevant variables is $C^{q}_{p-r}/C^{q}_{p} = 0.95$.
\end{example} \afterformula
With SRS (and $\alpha>0$), we ensure that more and more relevant variables are given to the tree
growing algorithm as iterations proceed and therefore we reduce the chance to
include totally useless trees in the ensemble. Note however that in finite
settings, there is a potential risk of overfitting when accumulating the
variables. The parameter $\alpha$ thus controls a new bias-variance tradeoff
and should be tuned appropriately. We will study the impact of SRS on accuracy
empirically in Section \ref{sec:empirical}.

%\vspace{-5mm}
\beforesection
\section{Theoretical analysis}
\label{sec:analysis}
\aftersection
In this section, we carry out a theoretical analysis of the proposed method
when seen as a feature selection technique. This analysis is performed in
asymptotic sample size condition and assuming that all features are
discrete. We proceed in two steps. First, we study the soundness of the
algorithm, ie., its capacity to retrieve the relevant variables when the number
of trees is infinite. Second, we study its convergence properties, ie. the
number of trees needed to retrieve all relevant variables in different
scenarios.

\beforesubsection
\subsection{Soundness}
\label{sec:soundness}
\aftersubsection

Our goal in this section is to characterize the sets of features
$F^{K,\alpha}_{q,\infty}$ that are identified by the SRS algorithm, depending
on the value of its parameters $q$, $\alpha$, and $K$, in an asymptotic
setting, ie. assuming an infinite sample size and an infinite forest
($T=\infty$). Note that in asymptotic setting, a variable is relevant as soon
as its importance in one of the tree is strictly greater than zero and we thus
have the following equivalence for all variables $X\in V$:
$$X\in F^{K,\alpha}_{q,\infty} \Leftrightarrow Imp^{K,\alpha}_{q,\infty}(X)>0$$
Furthermore, in infinite sample size setting, irrelevant variables always get a
zero importance and thus, whatever the parameters, we have the
following property for all $X\in V$:
$$X\mbox{ irrelevant} \Rightarrow X\notin F^{K,\alpha}_{q,\infty} \mbox{ (and }
Imp^{K,\alpha}_{q,\infty}(X)=0\mbox{)}.$$ The method parameters thus only affect
the number and nature of the relevant variables that can be found. Denoting by $r$ ($\leq p$) the
number of relevant variables, we will analyse separately the case $r\leq
q$ (all relevant variables can fit into memory) and the case $r>q$ (all relevant
variables can not fit into memory).

\beforeparagraph\paragraph{All relevant variables can fit into memory ($r\leq q$).}

Let us first consider the case of the RS method ($\alpha=0$). In this case,
\citet{louppe13-nips} have shown the following asymptotic formula for the
importances computed with totally randomized trees ($K=1$):
\begin{eqnarray}
Imp_{q,\infty}^{1,0}(X) = \sum_{k=0}^{q-1} \dfrac{1}{C_p^k} \sum_{B\in {\cal
    P}_k(V^{-m})} I(X;Y|B),
\end{eqnarray} 
where ${\cal P}_k(V^{-m})$ is the set of subsets of $V^{-m}=V\setminus \{x_m\}$
of cardinality $k$. Given that all terms are positive, this sum will be
strictly greater than zero if and only if there exists a subset $B\subseteq
V$ of size at most $q-1$ such that $Y\nindep X|B$, or equivalently if $deg(X)<
q$. When $\alpha=0$, RS with $K=1$ will thus find all and only the
relevant variables of degree at most $q-1$. Given Proposition
\ref{prop:only-relevant}, the degree of a variable $X$ can not be larger than
$r-1$ and thus as soon as $r\leq q$, we have the guarantee that RS with $K=1$
will find all and only the relevant variables. Actually, this result remains
valid when $\alpha>0$. Indeed, asymptotically, only relevant variables
will be selected in the $F$ subset by SRS and given that all relevant variables
can fit into memory, cumulating them will not impact the ability of SRS to
explore all conditioning subsets $B$ composed of relevant variables. We thus
have the following result:
\begin{proposition}
  $\forall \alpha$, if $r\leq q$: \quad $X \in F_{q,\infty}^{1,\alpha} \mbox{ iff }X\mbox{ is relevant}.$
\end{proposition} \afterformula

In the case of non-totally randomized trees ($K>1$), we lose the guarantee to
find all relevant variables even when $r\leq q$. Indeed, there is potentially a
masking effect due to $K>1$ that might prevent the conditioning needed for a
given variable to be relevant to appear in a tree branch.  However, we have the
following general result:
\begin{theorem}\label{th:stronglyrelpruned}
  $\forall \alpha, K$, if $r\leq q$:\quad
$X\mbox{ strongly relevant} \Rightarrow X \in F_{q,\infty}^{K,\alpha}$ \quad(Proof in Appendix~\ref{app:stronglyrelpruned})
\end{theorem} \afterformula
% \begin{proof} See .
% \end{proof}\vspace{-1em}
%\vspace{-0.4cm}
There is thus no masking effect possible for the strongly relevant features
when $K>1$ as soon as the number of relevant features is lower than $q$. For a
given $K$, the features found by SRS will thus include all strongly relevant
variables and some (when $K>1$) or all (when $K=1$) weakly relevant ones. It is easy to
show that increasing $K$ can only decrease the number of weakly relevant
variables found. Using $K=1$ will thus provide a solution for the {\bf all-relevant}
problem, while increasing $K$ will provide a better and better approximation of
the {\bf minimal-optimal} problem in the case of strictly positive distributions (see
Section \ref{sec:relevance}).

Interestingly, Theorem \ref{th:stronglyrelpruned} remains true when $q=p$, ie.,
when forests are grown without any feature sampling. It thus extends Theorem
\ref{th:louppe} from \cite{louppe13-nips} for arbitrary $K$ in the case of
standard random forests.

%% Discuss: the more randomization, the more weakly relevant variables will be
%% include into F. One should thus use high K to filter out weakly relevant
%% variable or to find a better approximation to the minimal ... when the
%% distribution is strictly positive. We will also discuss the impact of alpha and
%% K on convergence in the next section.

\beforeparagraph\paragraph{All relevant variables can not fit into memory ($r>q$).}

%degree(q) si K=1 et alpha=0. 

When all relevant variables can not fit into memory, we do not have the
guarantee anymore to explore all minimal conditionings required to find all
(strongly or not) relevant variables, whatever the values of $K$ and
$\alpha$. When $\alpha=0$, we have the guarantee however to identify the
relevant variables of degree strictly lower than $q$. When $\alpha>1$, some
space in memory will be devoted to previously found variables that will
introduce some further masking effect. We nevertheless have the following
general results (without proof):
\begin{proposition} $\forall X:$
  \quad$X\mbox{ relevant and } deg(X)<(1-\alpha)q \Rightarrow X\in F^{1,\alpha}_{q,\infty}.$
\end{proposition}
\begin{proposition} $\forall K,X:$
\quad$X\mbox{ strongly relevant and } deg(X)<(1-\alpha)q \Rightarrow X\in F^{K,\alpha}_{q,\infty}.$
\end{proposition}
\afterformula
%\pgnote{Il faudrait v\'erifier que la proposition 6 est bien correcte.\\}
In these propositions, $(1-\alpha)q$ is simply the amount of memory that
always remains available for the exploration of variables not yet found
relevant.

\beforeparagraph\paragraph{Discussion.}

Results in this section show that SRS is a sound approach for feature selection
as soon as either the memory is large enough to contain all relevant variables
or the degree of the relevant variables is not too high. In this latter case,
the approach will be able to detect all strongly relevant variables whatever
its parameters ($K$ and $\alpha$) and the total number of features $p$. Of
course, these parameters will have a potentially strong influence on the number
of trees needed to reach convergence (see the next section) and the performance
in finite setting.

\beforesubsection
\subsection{Convergence}\label{sec:convergence}
\aftersubsection

Results in the previous section show that accumulating relevant variables has
no impact on the capacity at finding relevant variables asymptotically (when
$r\leq q$). It has however a potentially strong impact on the convergence speed
of the algorithm, as measured for example by the expected number of trees
needed to find all relevant variables. Indeed, when $\alpha=0$ and $q\ll p$,
the number of iterations/trees needed to find relevant variables of high degree
can be huge as finding them requires to sample them together with all features
in their conditioning. Given Proposition 2, we know that a minimum subset $B$
such that $X\nindep Y|B$ for a relevant variable $X$ contains only relevant
variables. This suggests that accumulating previously found relevant features
can improve significantly the convergence, as each time one relevant variable
is found it increases the chance to find a relevant variable of higher degree
that depends on it. In what follows, we will quantify the effect of
accumulation on convergence speed in different best-case and worst-case
scenarios and under some simplifications of the tree building procedure. We
will conclude by a theorem highlighting the interest of the SRS method in the
general class of PC distributions.

\beforeparagraph\paragraph{Scenarios and assumptions.}
%% more interesting to consider specific scenarios (including the best-case and
%% the worst-case) to show the real improvement capacity of the proposed
%% algorithm. With this in mind, we first perform combinatorial analyses to
%% compute the expected number of iteratons needed to find all relevant
%% variables in the different scenarios. We then perform numerical simulations
%% to analyze more general settings for these scenarios.
The convergence speed is in general very much dependent on the data
distribution. We will study here the following three specific scenarios (where
features $\{X_1,\ldots,X_r\}$ are the only relevant features):\vspace{-0.5em}
\begin{itemize}[leftmargin=*]
\item{\bf Chaining:} The only and minimal conditioning that makes variable
  $X_i$ relevant is $\{X_1,\ldots,X_{i-1}\}$ (for $i=1,\ldots,r$). We
  thus have $deg(X_i)=i-1$. This scenario should correspond to the most
  favorable situation for the SRS algorithm.\vspace{-0.5em}
\item{\bf Clique:} The only and minimal conditioning that makes variable
  $X_i$ relevant is $\{X_1,\ldots,X_{i-1},X_{i+1},\ldots,X_r\}$ (for
  $i=1,\ldots,r$). We thus have $deg(X_i)=r-1$ for all $i$. This is a rather
  defavorable case for both RS and SRS since finding a relevant variable implies
  to draw all of them at the same iteration. \vspace{-0.5em}%%  However, let us note that we expect
  %% SRS to be faster in finding all features since the search will be facilitated
  %% (only $r-1$) after finding the first one while RS faces the same difficulty to
  %% find all features.
\item{\bf Marginal-only:} All variables are marginally relevant. We will
  furthermore make the assumption that these variables are all strongly
  relevant. They can not be masked mutually.  This scenario is the most
  defavorable case for SRS (versus RS) since accumulating relevant variables is
  totally useless to find the other relevant variables and it should actually
  slow down the convergence as it will reduce the amount of memory left for
  exploration.
\end{itemize}
\vspace{-0.4em} In Appendix \ref{sec:avgtime}, we provide explicit
formulation of the expected number of iterations needed to find all
$r$ relevant features in the chaining and clique scenarios both when
$\alpha=0$ (RS) and $\alpha=1$ (SRS). In Appendix \ref{sec:mc}, we
provide order 1 Markov chains that model the evolution through the
iterations of the number of variables found in the three scenarios
when $\alpha=0$ and $\alpha=1$. These chains can be used to compute
numerically the expected number of relevant variables found through
the iterations (and in the case of the marginal-only setting, the
expected number of iterations to find all variables).  These
derivations are obtained assuming $r\leq q$, $K=q$, and under
additional simplifying assumptions detailed in Appendix~\ref{app:assumption-convergence}.

\begin{figure*}[htb]
  \centerline{
  \includegraphics[width=0.28\linewidth]{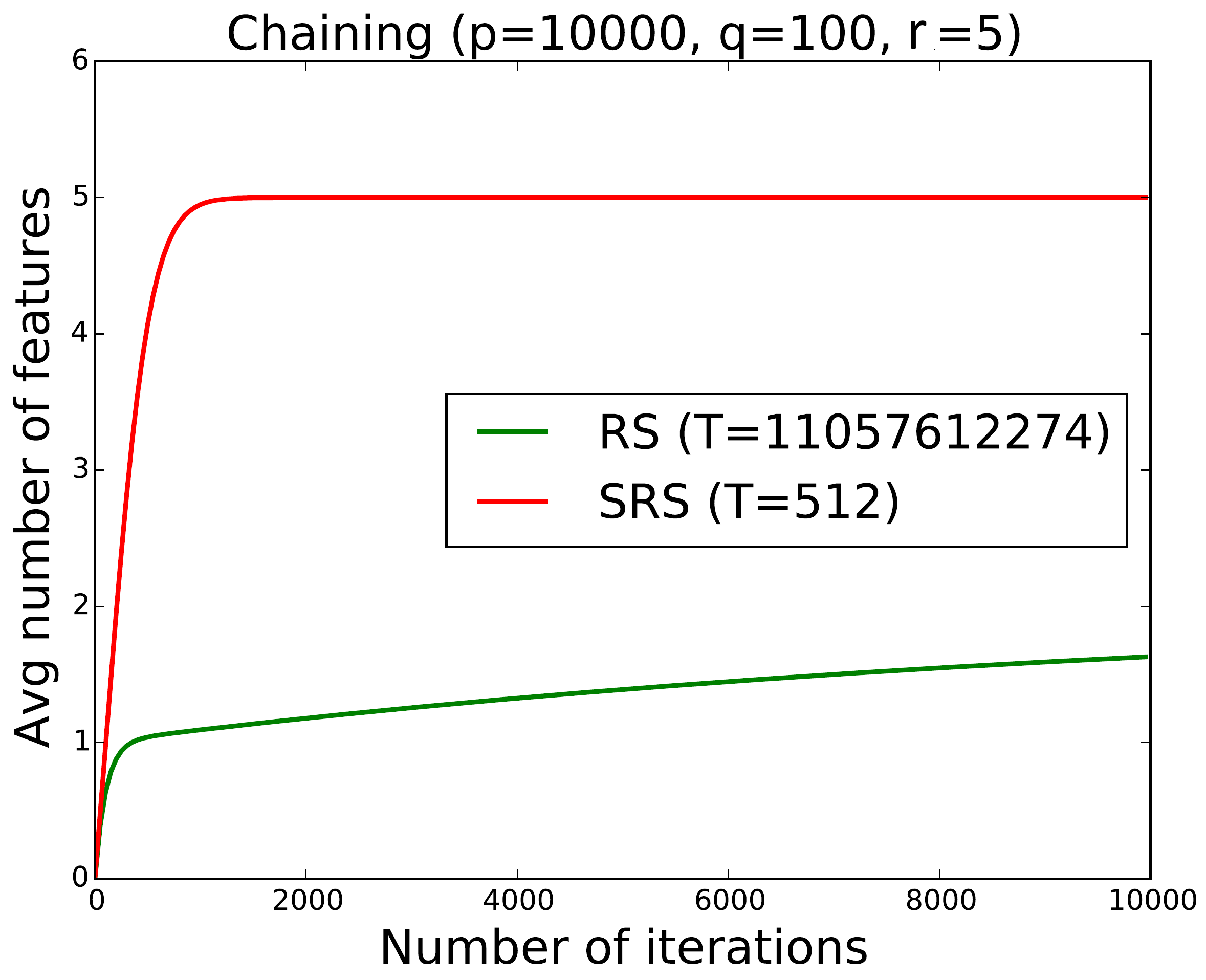}
  \includegraphics[width=0.28\linewidth]{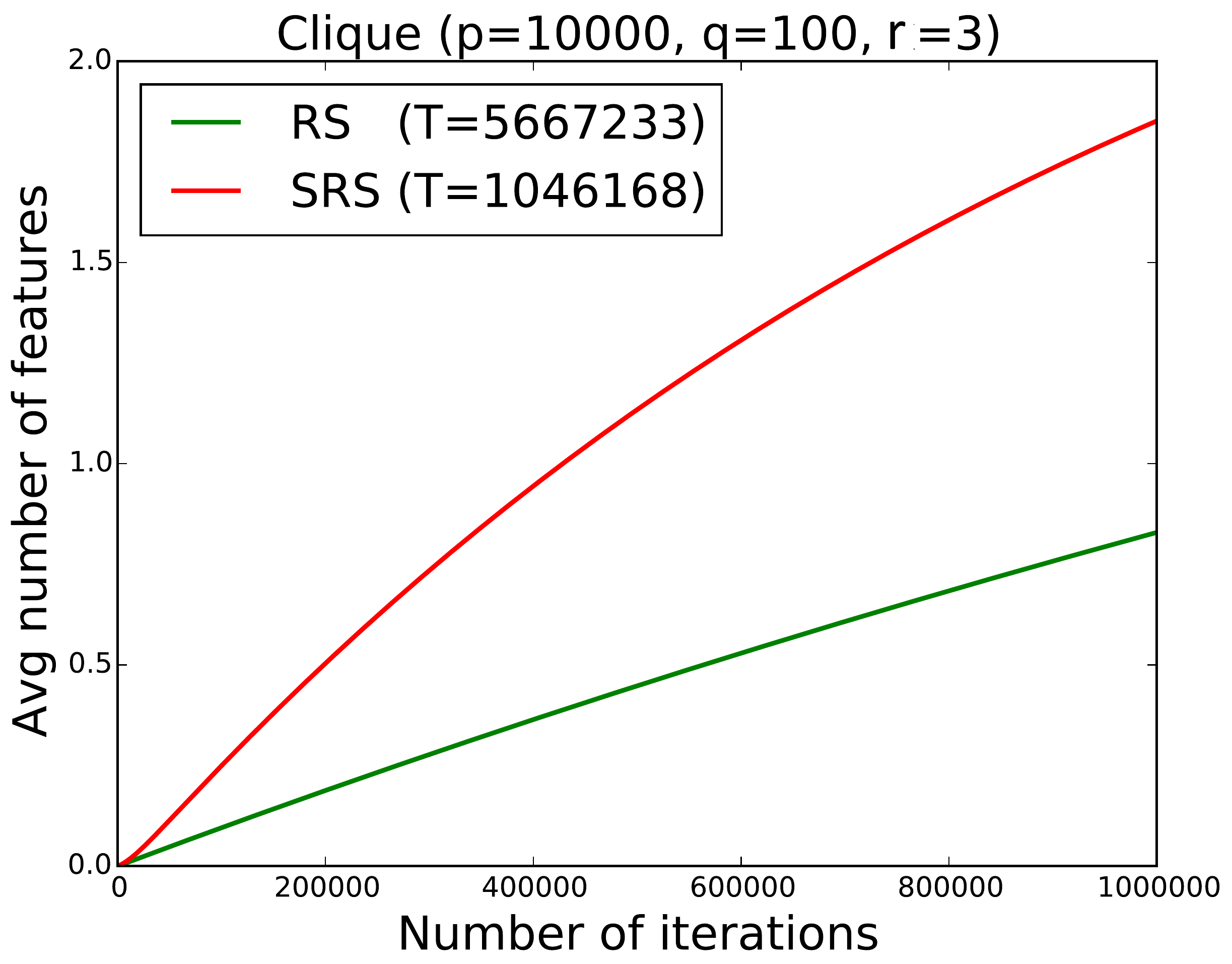}
  \includegraphics[width=0.28\linewidth]{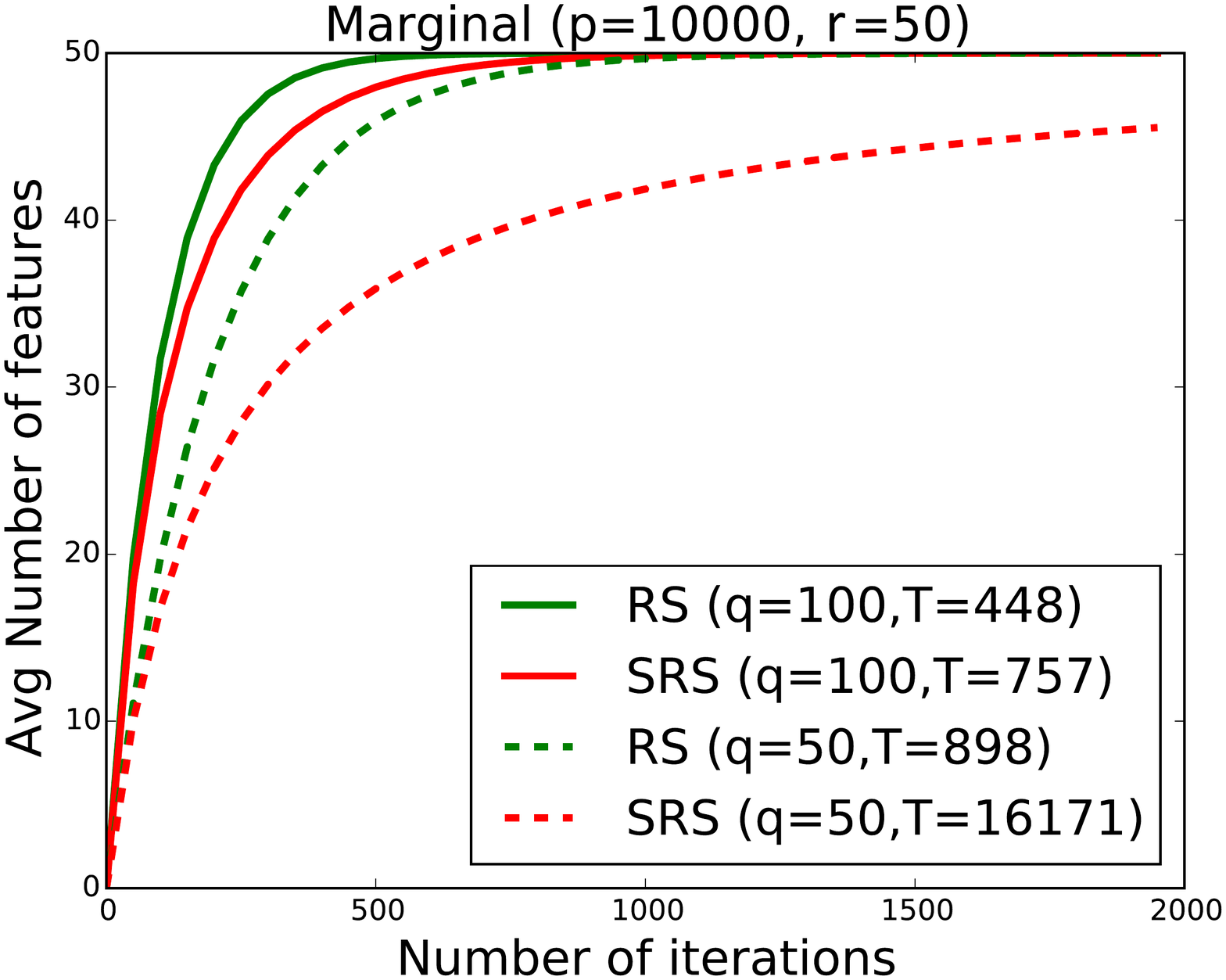}}
  \caption{Evolution of the number of selected features in the different scenarios.\label{fig:evolution}}
  \vspace{-1em}
\end{figure*}

\beforeparagraph\paragraph{Results and discussion.}
%% Using results in Appendices~\ref{sec:avgtime} and \ref{sec:mc}, we can perform
%% some interesting simulations and extract some numerical results.
Tables \ref{tab:timechaining}, \ref{tab:timeclique}, and \ref{tab:timemarginal}
show the expected number of iterations needed to find all relevant variables
for various configurations of the parameters $p$, $q$, and $r$, in the three
scenarios. Figure \ref{fig:evolution} plots the expected number of variables
found at each iteration both for RS and SRS in the three scenarios for some
particular values of the parameters.

\begin{table*}[htb]
\tiny
\centering
\caption{Expected number of iterations needed to find all relevant variables
  for various configurations of parameters $p$, $q$ and $r$ with RS
  ($\alpha=0$) and SRS ($\alpha=1$) in the three scenarios.}
% \vspace{1em}
%% \subfloat[Chaining.\label{tab:timechaining}]{
%%   \begin{tabular}{l|cc}
%%     Config & RS & SRS\\
%%     \hline
%%     $p=10^4%10000
%%     ,q=100,r=1$ & 100 & 100\\
%%     $p=10^4%10000
%%     ,q=100,r=2$ & 10100 & 200\\
%%     $p=10^4%10000
%%     ,q=100,r=3$ & $>10^6$ & 301\\
%%     $p=10^4%10000
%%     ,q=100,r=5$ & $>10^{10}$ & 506\\
%%     $p=10^5%100000
%%     ,q=100,r=3$ & $>10^9$ & 3028\\
%%   \end{tabular} 
%% }
%% \hspace{1.5em}
%% \subfloat[Clique.\label{tab:timeclique}]{
%%   \begin{tabular}{l|cc}
%%     Config & RS & SRS\\
%%     \hline
%%     $p=10^4,q=100,r=1$ & 100 & 100\\
%%     $p=10^4,q=100,r=2$ & 30300 & 10302\\
%%     $p=10^4,q=100,r=3$ & $5\cdot 10^6$ & $10^6$\\
%%     $p=10^4,q=100,r=4$ & $9\cdot  10^{8}$ & $10^8$\\
%%     $p=10^4,q=10^3,r=4$ & $83785$ & $11635$\\
%%   \end{tabular}  
%% }
%% \hspace{1.5em}
%% \subfloat[Marginal-only.\label{tab:timemarginal}]{  
%% \begin{tabular}{l|cc}
%%     Config & RS & SRS\\
%%     \hline
%%     $p=10^4%10000
%%     ,q=100,r=10$ & 291 & 312\\
%%     $p=10^4%10000
%%     ,q=100,r=50$ & 448 & 757\\
%%     $p=10^4%10000
%%     ,q=100,r=90$ & 506 & 2797\\
%%     $p=10^4%10000
%%     ,q=100,r=100$ & 1123 & 16187\\
%%     $p=25000
%%     ,q=100,r=50$ & 1123 & 1900\\
%%   \end{tabular}
%% } \vspace{-1em}

\subfloat[Chaining.\label{tab:timechaining}]{
  \begin{tabular}{l|cc}
    Config (p,q,r) & RS & SRS\\
    \hline
    $10^4%10000
    ,100,1$ & 100 & 100\\
    $10^4%10000
    ,100,2$ & 10100 & 200\\
    $10^4%10000
    ,100,3$ & $>10^6$ & 301\\
    $10^4%10000
    ,100,5$ & $>10^{10}$ & 506\\
    $10^5%100000
    ,100,3$ & $>10^9$ & 3028\\
  \end{tabular} 
}
\hspace{1.5em}
\subfloat[Clique.\label{tab:timeclique}]{
  \begin{tabular}{l|cc}
    Config (p,q,r) & RS & SRS\\
    \hline
    $10^4,100,1$ & 100 & 100\\
    $10^4,100,2$ & 30300 & 10302\\
    $10^4,100,3$ & $5\cdot 10^6$ & $10^6$\\
    $10^4,100,4$ & $9\cdot  10^{8}$ & $10^8$\\
    $10^4,10^3,4$ & $83785$ & $11635$\\
  \end{tabular}  
}
\hspace{1.5em}
\subfloat[Marginal-only.\label{tab:timemarginal}]{  
\begin{tabular}{l|cc}
    Config (p,q,r) & RS & SRS\\
    \hline
    $10^4%10000
    ,100,10$ & 291 & 312\\
    $10^4%10000
    ,100,50$ & 448 & 757\\
    $10^4%10000
    ,100,90$ & 506 & 2797\\
    $10^4%10000
    ,100,100$ & 1123 & 16187\\
    $25000
    ,100,50$ & 1123 & 1900\\
  \end{tabular}
} \vspace{-1em}
\end{table*}

From these results, we can draw several conclusions. In all cases, expected times (ie., number of iterations/trees to find all relevant
variables) depend mostly on the ratio $\frac{q}{p}$, not on absolute values of
$q$ and $p$. The larger this ratio, the faster the convergence. Parameter $r$
has a strong impact on convergence speed in all three scenarios.

The most impressive improvements with SRS are obtained in the {\bf chaining}
hypothesis, where convergence is improved by several orders of magnitude (Table
\ref{tab:timechaining} and Figure \ref{fig:evolution}a) . At fixed $p$ and $q$,
the time needed by RS indeed grows exponentially with $r$ ($\simeq
(\frac{p}{q})^r$ if $r\ll q$), while time grows linearly with $r$ for the SRS
method ($\simeq r\frac{p}{q}$ if $r\ll q$) (see Eq.  (\ref{ref:naive}) and
(\ref{ref:smart}) in Appendix \ref{sec:avgtime}).

In the case of {\bf cliques}, both RS and SRS need many iterations to find all
features from the clique (see Table \ref{tab:timeclique} and Figure
\ref{fig:evolution}b). SRS goes faster than RS but the improvement is not as
important as in the chaining scenario. This can be explained by the fact that
SRS can only improve the speed when the first feature of the clique has been
found. Since the number of iterations needed to find the $r$ features from the
clique for RS is close to $r$ times the number of iterations needed to find one
feature from the clique, SRS can only decrease at best the number of iterations
by approximately a factor $r$ (see Eq.  (\ref{eqn:naiveclique}) and
(\ref{eqn:smartclique}) in Appendix \ref{sec:avgtime}).
  
In the {\bf marginal-only} setting, SRS is actually slower than RS because the
only effect of cumulating the variables is to leave less space in memory for
exploration. The decrease of computing times is however contained when $r$ is
not too close to $q$ (see Table \ref{tab:timemarginal} and Figure
\ref{fig:evolution}c).

Since we can obtain very significant improvement in the case of the chaining
and clique scenarios and we only increase moderately the number of iterations
in the marginal-only scenario (when $r$ is not too close from $q$), we can
reasonably expect improvement in general settings that mix these scenarios.
% \vspace{-1mm}
\beforeparagraph\paragraph{PC distributions and chaining.}

Chaining is the most interesting scenario in terms of convergence improvement
through variable accumulation. In this scenario, SRS makes it possible to find
high degree relevant variables with a reasonable amount of trees, when finding
these variables would be mostly untractable for RS. We provide below two
theorems that show the practical relevance of this scenario in the specific
case of PC distributions.

A PC distribution is defined as a strictly positive (P) distribution that
satisfies the composition (C) property stated as follows
\cite{nilsson2007consistent}:
\begin{property}
  For any disjoint sets of variables $R , S, T, U \subseteq V\cup\{Y\}$:
  $$S\indep T|R\mbox{ and }S\indep U|R \Rightarrow S\indep T\cup U | R$$
\end{property} \afterformula
The composition property prevents the occurence of cliques and is preserved
under marginalization. PC actually represents a rather large class of
distributions that encompasses for example jointly Gaussian distributions and
DAG-faithful distributions \cite{nilsson2007consistent}.

The composition property allows to make Proposition \ref{prop:only-degree} more
stringent in the case of PC:
\begin{proposition} \label{prop:only-degree-pc}
Let $B$ denote a minimal subset $B$ such that $Y\nindep X|B$ for a relevant
variable $X$. If the distribution $P$ over $V\cup\{Y\}$ is PC, then for all
$X'\in B$, $deg(X')<|B|$. \quad (Proof in Appendix~\ref{app:only-degree-pc})
\end{proposition} \afterformula
% \vspace{-1em}
% \begin{proof}
%   See Appendix~\ref{app:only-degree-pc}.
%   \end{proof}\vspace{-1em}
In addition, one has the following result:
\begin{theorem}\label{th:pcchain}
  For any PC distribution, let us assume that there exists a non empty
  minimal subset $B=\{X_1,\ldots,X_k\}\subset V\setminus \{X\}$ of
  size $k$ such that $X\nindep Y |B$ for a relevant variable $X$.
  Then, variables $X_1$ to $X_k$ can be ordered into a sequence
  $\{X'_1,\ldots,X'_k\}$ such that $deg(X'_i)<i$ for all
  $i=1,\ldots,k$. \quad(Proof in Appendix~\ref{app:pcchain})
\end{theorem} \afterformula
% \vspace{-1em}
%   \begin{proof}
%  See Appendix~\ref{app:pcchain}.
%  \end{proof}\vspace{-1em}

 This theorem shows that, when the data distribution is PC, for all relevant
 variables of degree $k$, the $k$ variables in its minimal conditioning form a
 chain of variables of increasing degrees (at worst). For such distribution, we
 thus have the guarantee that SRS find all relevant variables
 with a number of iterations that grows almost only linearly with the maximum
 degree of relevant variables (see Eq.\ref{ref:smart} in Appendix
 \ref{sec:avgtime}), while RS would be unable to find relevant variables of
 even small degree.

\beforesection
\section{Experiments}\label{sec:empirical}
\aftersection

\begin{figure}[tb]
\centering
  \subfloat[SRS with $q=0.05 \times p $ on a dataset with $p=50000$ features and $r=20$ relevant features.]{\includegraphics[width=0.35\linewidth]{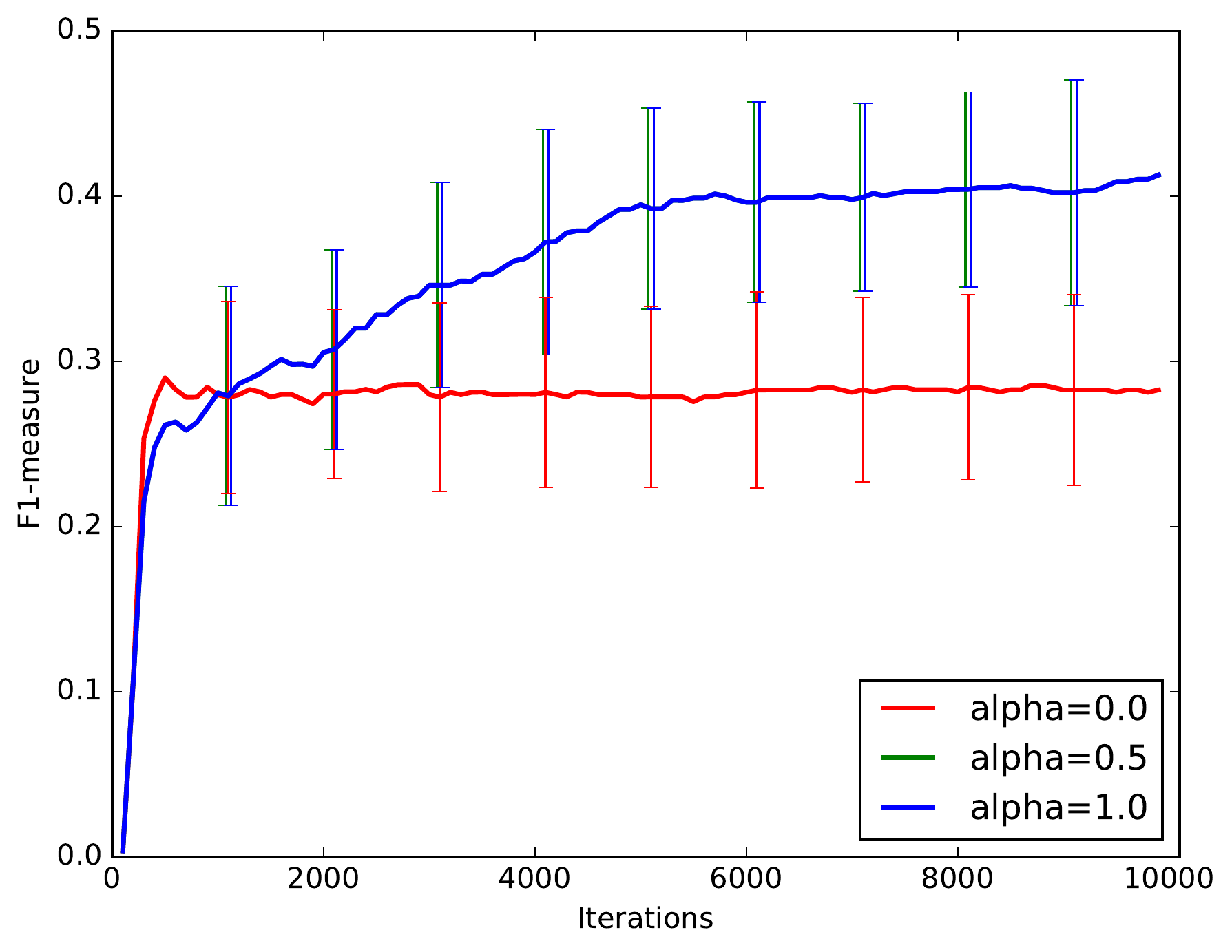}}\hspace{1em}
  \subfloat[SRS with $q=0.005 \times p$ on a dataset with $p=50000$ features and $r=20$ relevant features.]{\includegraphics[width=0.35\linewidth]{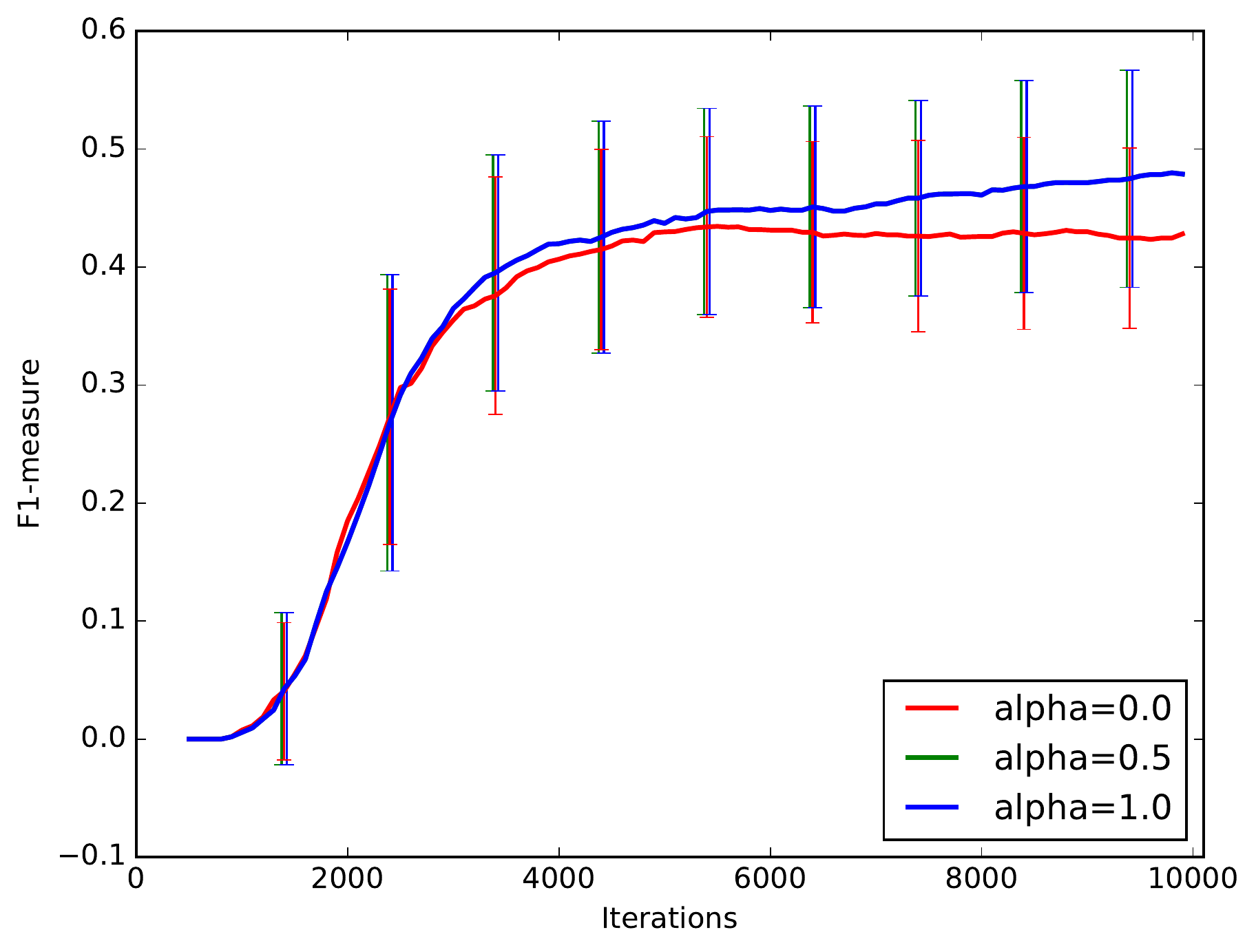}}
  \caption{Evolution of the evaluation of the feature subset found by RS and SRS using the F1-measure computed with respect to relevant features. A higher value means that more relevant features have been found. This experiment was computed on an artificial dataset (similar to madelon) of 50000 features with 20 relevant features and for two sizes of memory.}
  \label{fig:pr} \vspace{-0em}
\end{figure}

Although our main contribution is the theoretical analysis in
asymptotic setting of the previous section, we present here a few
preliminary experiments in finite setting as a first illustration of
the potential of the method.%%  A more extensive empirical analysis is
%% left as future work.
One of the main difficulties to implement the SRS algorithm as presented in
Algorithm \ref{algo:SRS} is step 2(c) that decides which variable should be
incorporated in $F$ at each iteration. In infinite sample size
setting, a variable with a non-zero importance in a single tree is guaranteed
to be truly relevant. Mutual informations estimated from finite samples however
will always be greater than 0 even for irrelevant variables. One should thus
replace step 2(c) by some statistical significance tests to avoid the
accumulation of irrelevant variables that would jeopardize the convergence of
the algorithm. In our experiments here, we use a random probe (ie., an
artificially created irrelevant variable) to derive a statistical measure
assessing the relevance of a variable \cite{stoppiglia03}. Details about this
test are given in Appendix~\ref{app:results}.

Figure \ref{fig:pr} evaluates the feature selection ability of SRS for three values
of $\alpha$ (including $\alpha=0$) and two memory sizes (250 and 2500) on an
artificial dataset with 50000 features, among which only 20 are relevant (see
Appendix~\ref{app:results} for more details). The two plots show the evolution
of the F1-score comparing the selected features (in $F$) with the truly
relevant ones as a function of the number of iterations. As expected, SRS
($\alpha>0$) is able to find better feature subsets than RS ($\alpha=0$) for
both memory sizes and both values of $\alpha>0$.

Additional results are provided in Appendix~\ref{app:results} that compare the
accuracy of ensembles grown with SRS for different values of $\alpha$ and on 13
classification problems. These comparisons clearly show that accumulating the
relevant variables is beneficial most of the time (eg., SRS with $\alpha=0.5$
is significantly better than RS on 7 datasets, comparable on 5, and
significantly worse on only 1). Interestingly, SRS ensembles with $\alpha=0.5$ are also most of
the time significantly better than ensembles of trees grown without memory
constraint (see Appendix~\ref{app:results} for more details).

\beforesection
\section{Conclusions and future work} 
\aftersection

Our main contribution is a theoretical analysis of the SRS (and RS) methods in
infinite sample setting. This analysis showed that both methods provide some
guarantees to identify all relevant (or all strongly) relevant variables as
soon as the number of relevant variables or their degree is not too high with
respect to the memory size. Compared to RS, SRS can reduce very strongly the
number of iterations needed to find high degree variables in particular in the
case of PC distributions. We believe that our results shed some new light on random
subspace methods for feature selection in general as well as on tree-based
methods, which should help designing better feature selection procedures.

Although some preliminary experiments were provided that support the
theoretical analysis, more work is clearly needed to evaluate the approach
empirically on controlled and real high-dimensional problems. We believe
that the statistical test used to decide which feature to include in the
relevant set should be improved with respect to our first implementation based
on the introduction of a random probe. One drawback of the SRS method with
respect to RS is that it can not be parallelized anymore because of its
sequential nature. It would be interesting to design and study variants of the
method that are allowed to grow parallel ensembles at each iteration instead of
single trees. Finally, relaxing the main hypotheses of our theoretical analysis
would be also of course of great interest.

% \newpage \newpage
% \section*{Acknowledgements} 
% Antonio Sutera is a recipient of a FRIA grant from the FNRS (Belgium) and
% acknowledges its financial support. This work is supported by the IUAP DYSCO, initiated by the Belgian State, Science Policy Office.  Computational resources have been provided by the Consortium
% des Equipements de Calcul Intensif (CECI), funded by the Fonds de la Recherche Scientifique de Belgique (F.R.S.-FNRS).
\newpage
\bibliography{references}
% \bibliographystyle{natbib}

% \vfill


\section*{Supplementary material}

# Random Subspace with Trees for Feature Selection Under Memory Constraints
*Supplementary materials*

**Antonio Sutera**
University of Liège, Belgium
sutera.antonio@gmail.com

**Célia Châtel**
Aix-Marseille University, France

**Gilles Louppe**
New York University, USA

**Louis Wehenkel**
University of Liège, Belgium

**Pierre Geurts**
University of Liège, Belgium

## A  Proof of Proposition 1

**Proposition 1.** *A minimal subset $B$ such that $Y \not\perp\!\!\!\perp X|B$ for a relevant variable $X$ contains only relevant variables.*

*Proof.* Let us assume that $B$ contains an irrelevant variable $X_i$. Let us denote by $B^{-i}$ the subset $B \setminus \{X_i\}$. Since $X_i$ is irrelevant, we have $Y \perp\!\!\!\perp X_i|B^{-i}\cup\{X\}$. Given that $B$ is minimal we furthermore have $Y \perp\!\!\!\perp X|B^{-i}$ where $B^{-i} = B \setminus \{X_i\}$. By using the contraction property of any probability distribution **?**, one can then conclude from these two independences that $Y \perp\!\!\!\perp \{X, X_i\}|B^{-i}$ and, by using the weak union property, that $Y \perp\!\!\!\perp X|B$, which proves the theorem by contradiction. □

## B  Proof of Proposition 2

**Proposition 2.** *Let $B$ denote a minimal subset such that $Y \not\perp\!\!\!\perp X|B$ for a relevant variable $X$. For all $X' \in B$, $deg(X') \leq |B|$.*

*Proof.* If we reduce the set of features $V$ to a new set $V' = B \cup \{X\}$, $X$ will remain relevant, as well as all features in $B$, given Proposition 1. So, for any feature $X'$ in $B$, there exists a subset $B' = B \cup \{X\} \setminus X'$ such that $Y \not\perp\!\!\!\perp X'|B'$ and the degree of $X'$ is therefore $\leq |B|$. □

## C  Proof of Theorem 2

**Theorem 2.** $\forall \alpha, K$, *if* $r \leq q$:

$$X \text{ strongly relevant} \Rightarrow X \in F_{q,\infty}^{K,\alpha}$$

*Proof.* Let us first consider the case $\alpha = 0$ (i.e., the RS method).

Since $X$ is strongly relevant, there is at least one assignment of values to all variables but $X$ such that conditionally to this assignment, $Y$ is dependent on $X$. Let us consider all possible branches of length $q$ that can be generated that are compatible with this assignment. Let us show that there is at least one tree where $X$ was splitted along one of these branches with a non zero importance score.

All splits on irrelevant variables have zero score and thus every time we split on such variable, one could have instead splitted the node on a relevant variable because of randomization (due either to $K < q$ or due to the fact that when several variables have a zero score at some node, we assume

that one of them is picked at random to split), except if all relevant variable have been tested above that node. This means that whatever $K$, all branches that could be constructed using a fully grown tree (using all features) with the $r$ relevant features can be reduced to a branch of length $r$ with only relevant features including variable $X$. When $K < q$, randomization ensures that there is a non zero probability that feature $X$ is seen at the end of the branch.

Let us show that this also the case with $K = q$ by assuming that feature $X$ does not get a positive score when using $K = q$ and showing that this leads to a contradiction. If $X$ was splitted at the bottom of the branch, it would get a positive score since it is a strongly relevant feature. Therefore, a zero score means that $X$ is always splitted in the middle of the branch and that all other features also had a zero score when $X$ was selected. It thus means that by chance another feature could have been used instead of $X$ and $X$ has thus a non zero probability of being splitted at least one level below in another tree. Applying this argument recusively, it means that in this case, there is at least one tree where $X$ is splitted at the bottom of the branch corresponding to the assignment, which contradicts the fact that $X$ is splitted in the middle of the branch. $X$ thus always gets a non zero importance whatever $K$.

The generalization of this result to $\alpha > 0$ is straightforward. Given that only relevant variables can be kept in memory in case of $\alpha > 0$ and given that $q \geq r$, accumulating relevant variables does not prevent exploring all the branches required to find $X$.

□

## D  Convergence analysis

### D.1  Simplifying assumptions

Below, we compute analytically the average number of trees needed to find all relevant variables in the chaining and clique scenarios and we derive transition matrices of Markov chains that model the evolution of the number of variables found through the iterations in the three scenarios. These results are obtained assuming $K = q$ and $r \leq q$, and with either $\alpha = 0$ (RS) or $\alpha = 1$ (SRS).

To make these derivations possible and independent of a particular data distribution, one needs furthermore to simplify the decision tree growing algorithm in the case of the chaining and clique scenarios. In what follows, trees are thus assumed to be grown such that a unique variable is selected at each tree level and this variable is selected at random among all variables $X$ such that $Y \not\perp\!\!\!\perp X | B$ where $B$ is the set of all variables tested at previous levels.

In the clique scenario, this assumption implies that only one variable of the clique will get a non-zero importance when all clique variables are selected at one iteration of RS/SRS (since only the last variable of the clique tested along a tree branch can get a non-zero score and this variable is the same in each branch given our tree growing assumption). This corresponds to a pessimistic scenario. Indeed, with standard unconstrained trees, several relevant variables could be found at one iteration given that the ordering of the variables, and thus the last variable of the clique tested, might differ from one tree branch to another. As a consequence, the tree growing assumption will lead to an overestimation of the number of trees needed to reach convergence. In the chaining scenario, the simplified tree growing algorithm implies that all relevant variables selected at one iteration of RS/SRS together with their minimal conditioning will get a non-zero importance. This corresponds this time to an optimistic scenario, as, with unconstrained trees, such variable might not be detected at one iteration depending on the exact data distribution. This will thus lead this time to an underestimation of the number of trees needed to reach convergence. Note however that, in both cases, these over/under-estimations will affect both RS and SRS in the same proportion and thus our assumption will not impact their relative performance.

Note that in the marginal-only scenario, given that all relevant variables are marginally and strongly relevant, they will always get a non-zero importance as soon as they are selected at one iteration. Our estimations below are thus not impacted by the simplification of the tree growing algorithm.

### D.2 Average times

**Chaining.** Let us denote by $T^{RS}_{chain}(i,p,q)$ ($1 \leq i \leq r$) the average number of iterations needed to find the feature $X_i$ of degree $i-1$ and by $T^{SRS}_{chain}(i,p,q)$ the average number of iterations needed to find the same feature with the SRS algorithm (that forces the selection of already found relevant variables). Given our assumptions above, each tree will be able to identify all relevant variables $X$ it gets as soon as it gets also the relevant variables in its minimal conditioning. Note that $T^{RS/SRS}_{chain}(i,p,q)$ can also be interpreted as the average time needed to find the first $i$ relevant features, given that one can not find $X_i$ without finding all features $X_j$ with $1 \leq j < i$. $T^{RS/SRS}_{chain}(r,p,q)$ also represents the average number of iterations needed to find all relevant variables under the chain assumption.

**Theorem 4.** *Under our assumptions, the $T^{RS}_{chain}$ function can be computed as follows:*

$$T^{RS}_{chain}(i,p,q) = \prod_{l=0}^{i-1} \frac{p-l}{q-l} \tag{1}$$

*Proof.* Indeed, $T^{RS}_{chain}(i,p,q)$ is the mean of a geometric distributed random variable with a probability of success defined as the probability of drawing the $i-1$ variables in $X_i$'s conditioning and $X_i$ at the same time, which is given by:

$$\frac{\binom{p-i}{q-i}}{\binom{p}{q}} = \prod_{l=0}^{i-1} \frac{q-l}{p-l}. \tag{2}$$

□

**Theorem 5.** *Under the same assumption, $T^{SRS}_{chain}(i,p,q)$ can be computed as follows:*

$$T^{SRS}_{chain}(i,p,q) = \sum_{l=0}^{i-1} \frac{p-l}{q-l} - (i-1) \tag{3}$$

*Proof.* Let us show this by induction on $i$. The base case corresponds to $i=1$. In this case, we have:

$$T^{SRS}_{chain}(1,p,q) = T^{RS}_{chain}(1,p,q) = \frac{p}{q},$$

which satisfies Eqn (3). Let us assume that Eqn. (3) is satisfied for $i < i'$ and let us show that it is satisfied for $i = i'$. $T^{SRS}_{chain}(i',p,q)$ can be defined as follows:

$$T^{SRS}_{chain}(i',p,q) = \frac{q}{p} T^{SRS}_{chain}(i'-1, p-1, q-1) + \\ (1 - \frac{q}{p})(1 + T^{SRS}_{chain}(i',p,q)). \tag{4}$$

One can indeed distinguish two cases:

- $X_1$ is selected at the first iteration (this happens with probability $q/p$): the average time needed to find feature $X_{i'}$ of degree $i'-1$ then becomes the time needed to find a feature of degree $i'-2$ when one is allowed to draw $q-1$ features among $p-1$, which is $T^{SRS}_{chain}(i'-1, p-1, q-1)$

- $X_1$ is not selected at the first iteration (this happens with probability $1-q/p$): in this case, the first iteration is useless and thus the number of iterations needed will be $1 + T^{SRS}_{chain}(i',p,q)$.

Eqn. (4) can be used to compute $T^{SRS}_{chain}$ recursively:

$$T^{SRS}_{chain}(i',p,q) = T^{SRS}_{chain}(i'-1, p-1, q-1) + (\frac{p}{q} - 1). \tag{5}$$

Deriving Eqn. (3) from Eqn. (5) is then straightforward, which concludes the proof by induction. □

Eqn. (3) shows that the average time needed to find the $i$ first features is equal to the sum of the time needed to find all features individually minus the number of features. This last term takes into account the fact that by chance, one might find several features at once.

**Clique.** Let us denote by $T_{cl}^{RS}(i, p, q)$ and $T_{cl}^{SRS}(i, p, q)$, the average time needed to find $i$ features (among $r$) from the clique respectively with the RS and the SRS algorithm. Given our assumptions above, when the tree growing algorithm is given all $r$ relevant features, it will be able to identify one (and only one) feature from the clique at random. If it has already found $i$ features from the clique, the chance to get a new one, when all $r$ features are selected among the $q$ ones, will thus be $(r - i)/r$, i.e., the probability to test one of the $r - i$ not yet found features after all other $r$ features from the clique.

**Theorem 6.**

$$T_{cl}^{RS}(i, p, q) = \left(\prod_{l=0}^{r-1} \frac{p-l}{q-l}\right) \cdot \left(\sum_{l=0}^{i-1} \frac{r}{r-l}\right) \tag{6}$$

*Proof.* The first factor in Eqn.(6) is the inverse of the probability of selecting all $r$ relevant features at once. Each term of the sum in the second factor corresponds to the inverse of the probability of testing a new relevant variables, not yet found, at the bottom of the tree. As discussed above, this probability is $\frac{r-l}{r}$ when we have already found $l$ features from the clique. □

**Theorem 7.**

$$T_{cl}^{SRS}(i, p, q) = \sum_{l=0}^{i-1} \frac{r}{r-l} \prod_{m=l}^{r-1} \frac{p-m}{q-m} \tag{7}$$

*Proof.* Each term of the sum represents the average time needed to find a new clique feature given that we have already found $l$ features. This time is equal to one over the probability of finding a new feature when we have already found $l$ of them. This latter is the probability of selecting among $q$ the $r - l$ missing relevant features (i.e., $\prod_{m=l}^{r} \frac{q-m}{p-m}$) times the probability of testing one of the missing relevant features at the bottom of the tree (i.e., $(r - l)/r$). □

When $i = 1$, $T_{cl}^{SRS}(1, p, q) = T_{cl}^{RS}(1, p, q)$. Intuitively, it indeed takes the same time for the RS and the SRS algorithms to find the first relevant features. When $i$ increases however, the SRS algorithm becomes faster and faster than the RS algorithm. Indeed, the RS algorithm always needs to find all $r$ clique features, while the SRS one only needs to find the $r - i$ missing relevant features.

### D.3 Markov chain interpretation

Let us denote by $N_t^{X,Y}$ the number of variables found for $t$ iterations, with $X = c$, $X = g$, and $X = m$ respectively for the chain hypothesis, the clique hypothesis and the marginal only hypothesis (as defined in the first section of this document) and $Y = n$ and $Y = s$ respectively for the RS and SRS algorithms. All these random variables follow order 1 Markov chains. The transition probabilities are provided below for each chain (without proof), under the assumptions given in Section D.1.

**Chain hypothesis.**

$$P(N_t^{c,n} = l_1 | N_t^{c,n} = l_2) = \begin{cases} 0 & \text{if } l_1 < l_2 \\ \frac{\binom{p-r}{q-l_1}}{\binom{p}{q}} & \text{if } l_1 > l_2 \\ 1 - \sum_{i=l_2+1}^{r} \frac{\binom{p-r}{q-i}}{\binom{p}{q}} & \text{if } l_1 = l_2 \end{cases} \tag{8}$$

$$P(N_t^{c,s} = l_1 | N_t^{c,s} = l_2) = \begin{cases} 0 & \text{if } l_1 < l_2 \\ \frac{\binom{p-r}{q-l_1}}{\binom{p-l_2}{q-l_2}} & \text{if } l_1 > l_2 \\ 1 - \sum_{i=l_2+1}^{r} \frac{\binom{p-r}{q-i}}{\binom{p-l_2}{q-l_2}} & \text{if } l_1 = l_2 \end{cases} \tag{9}$$

4

**Clique hypothesis.**

$$P(N_t^{g,n} = l_1 | N_t^{g,n} = l_2) = \begin{cases} 0 & \text{if } l_1 < l_2 \\ 1 - \frac{\binom{p-r}{q-r}}{\binom{p}{q}} \frac{r-l_2}{r} & \text{if } l_1 = l_2 \\ \frac{\binom{p-r}{q-r}}{\binom{p}{q}} \frac{r-l_2}{r} & \text{if } l_1 = l_2 + 1 \\ 0 & \text{if } l_1 > l_2 + 1 \end{cases} \quad (10)$$

$$P(N_t^{g,s} = l_1 | N_t^{g,s} = l_2) = \begin{cases} 0 & \text{if } l_1 < l_2 \\ 1 - \frac{\binom{p-r}{q-r}}{\binom{p-l_2}{q-l_2}} \frac{r-l_2}{r} & \text{if } l_1 = l_2 \\ \frac{\binom{p-r}{q-r}}{\binom{p-l_2}{q-l_2}} \frac{r-l_2}{r} & \text{if } l_1 = l_2 + 1 \\ 0 & \text{if } l_1 > l_2 + 1 \end{cases} \quad (11)$$

**Marginal only hypothesis.**

$$P(N_t^{m,n} = l_1 | N_t^{m,n} = l_2) = \begin{cases} 0 & \text{if } l_1 < l_2 \\ \frac{\binom{r-l_2}{l_1-l_2}\binom{p-r+l_2}{q-l_1+l_2}}{\binom{p}{q}} & \text{if } l_1 > l_2 \\ \frac{\binom{p-r+l_2}{q}}{\binom{p}{q}} & \text{if } l_1 = l_2 \end{cases} \quad (12)$$

$$P(N_t^{m,s} = l_1 | N_t^{m,s} = l_2) = \begin{cases} 0 & \text{if } l_1 < l_2 \\ \frac{\binom{r-l_2}{l_1-l_2}\binom{p-r}{q-l_1}}{\binom{p-l_2}{q-l_2}} & \text{if } l_1 > l_2 \\ \frac{\binom{p-r}{q-l_2}}{\binom{p-l_2}{q-l_2}} & \text{if } l_1 = l_2 \end{cases} \quad (13)$$

## E  Proof of Proposition 6

**Proposition 6.** *Let $B$ denote a minimal subset $B$ such that $Y \not\perp\!\!\!\perp X | B$ for a relevant variable $X$. If the distribution $P$ over $V \cup \{Y\}$ is PC, then for all $X' \in B$, $deg(X') < |B|$.*

*Proof.* Proposition 2 proves that the degree of all features in $B$ is $\leq |B|$ in the general case. Let us assume that there exists a feature $X' \in B$ of degree $|B|$ in the case of PC distribution. Since this property remain true when the set of features $V$ is reduced to a subset $V' = B \cup \{X\}$, the minimal $B'$ of $X'$ can only be $(B \setminus \{X_i\}) \cup \{X\}$. We thus have the following two properties:

$$Y \perp\!\!\!\perp X | B \setminus \{X'\}$$
$$Y \perp\!\!\!\perp X' | B' \setminus \{X\},$$

because $B$ and $B'$ are minimal. Together, by the composition property, they should imply that

$$Y \perp\!\!\!\perp \{X, X_i\} | B \setminus \{X_i\},$$

which implies, by weak union: $Y \perp\!\!\!\perp X | B$, which contradicts the hypothesis. □

## F  Proof of Theorem 3

**Theorem 3.** *For any PC distribution, let us assume that there exists a non empty minimal subset $B = \{X_1, \ldots, X_k\} \subset V \setminus \{X\}$ of size $k$ such that $X \not\perp\!\!\!\perp Y | B$ for a relevant variable $X$. Then, variables $X_1$ to $X_k$ can be ordered into a sequence $\{X'_1, \ldots, X'_k\}$ such that $deg(X'_i) < i$ for all $i = 1, \ldots, k$.*

*Proof.* Let us denote by $\{X'_1, X'_2, \ldots, X'_k\}$ the variables in $B$ ordered according to their degree, ie., $deg(X'_i) \leq deg(X'_{i+1})$, for $i = 1, \ldots, k-1$. Let us show that $deg(X'_i) < i$ for all $i = 1, \ldots, k$. If this property is not true, then there exists at least one $X'_i \in B$ such that $deg(X'_i) \geq i$. Let us denote by $l$ the largest $i$ such that $deg(X_i) \geq i$. Using a similar argument as in the proof of Proposition 6,

5

| Dataset  | # samples | # features |
|----------|-----------|------------|
| arcene   | 100       | 10000      |
| breast2  | 295       | 24496      |
| cina0    | 16033     | 132        |
| isolet   | 7797      | 617        |
| madelon  | 2000      | 500        |
| marti0   | 500       | 1024       |
| reged0   | 500       | 999        |
| secom    | 1567      | 591        |
| mnist    | 70000     | 784        |
| mnist3v8 | 13966     | 784        |
| mnist4v9 | 13782     | 784        |
| sido0    | 12678     | 4932       |
| tis      | 13375     | 927        |

Table 1: Dataset specifications

there exists some minimal subset $B' \subseteq B \setminus \{X_l\}$ such that $Y \not\perp X_l | B'$. Given that $deg(X_l) \geq l$, this subset $B$ should contain $l$ variables or more from $B \setminus \{X_l\}$. It thus contains at least one variable $X_m$ with $l < m \leq k$, and this variable is such that $deg(X_m) < m$. Given Proposition 6, if $B'$ is minimal and contains $X_m$, then for a PC distribution, $deg(X_m)$ should be strictly smaller than $|B'| \geq l$, which contradicts the fact that $X_m$ is after $X_l$ in the ordering and proves the theorem. □

## G  Details for Section 5

In this section, we give more details about our practical implementation of SRS and performed experiments.

### G.1  On the use of a random probe to distinguish relevant features from irrelevant features.

As explained in Section 5, we add an artificial irrelevant feature in data as a random probe. By comparison with that probe of importances scores, one can distinguish relevant features (better than the probe) from irrelevant features. Through iterations, we can compute a *p-value* score which is the percentage of times a variable has been better than the probe. If the *p-value* is above a given threshold $\beta$ then the feature is likely relevant. Moreover, a variable has to be sampled more than $L$ times in $Q$ sets to insure that the *p-value* is reliable. Then at each iteration, the variables that satisfy the two criteria are added to $F$. In the following experiments, we choose arbitrarily $L = 10$ and $\beta = 95\%$.

### G.2  On the datasets and on the protocol

We evaluate the accuracy of all these methods on a list of both artificial and real classifications problems (all but madelon are real data) described in Table 1 and publicly available in the UCI machine learning repository **?**. For each dataset, we separate it into two random partitions of the same size (i.e., the same number of samples) to have a training set and a test set. There is no optimization of the parameters. For all datasets, the procedure was repeated 50 times, using the same random partitions between all methods. Following results are averages over those 50 runs.

### G.3  Detailed results

Table 2 is average accuracy scores obtained on all datasets for each method for some parameters. We consider different sizes of memory (i.e., parameter $q$) and different value for the parameter $\alpha$ for the SRS algorithm. This allows to consider every behaviour of the SRS algorithm : without memory ($\alpha = 0$) which is equivalent to the Random Subspace method, with a full memory ($\alpha = 1$) and a non-full memory ($\alpha = 0.5$). For both methods (RS and SRS), a single extra-tree is build at each iteration. The randomization parameter of the extra-tree is set to its maximal value (ie., all features). For the tree-based ensemble methods, we consider different values for the randomization parameter. This parameter reduces the ability to consider the whole dataset in once and in that it relates in a way to the size of the memory of SRS. We choose for that parameter values of 0.01, 0.1 and 1 corresponding to considering respectively 1%, 10%, 100% of all features at each node.

|  | SRS | | | | | | | | | Tree-based ensemble methods | | | | | |
|  | q=0.01 | | | q=0.05 | | | q=0.1 | | | RF | | | ET | | |
|  | $\alpha$ | | | | | | | | | Randomization parameter $K$ | | | | | |
|  | 0.0 | 0.5 | 1.0 | 0.0 | 0.5 | 1.0 | 0.0 | 0.5 | 1.0 | 0.01 | 0.1 | 1 | 0.01 | 0.1 | 1 |
| --- | --- | --- | --- | --- | --- | --- | --- | --- | --- | --- | --- | --- | --- | --- | --- |
| arcene | 0.743 | 0.717 | 0.717 | 0.743 | 0.743 | 0.743 | 0.732 | 0.732 | 0.732 | 0.717 | 0.706 | 0.678 | 0.739 | 0.729 | 0.701 |
| breast2 | 0.649 | 0.647 | 0.647 | 0.651 | 0.651 | 0.650 | 0.654 | 0.654 | 0.654 | 0.646 | 0.649 | 0.649 | 0.650 | 0.654 | 0.651 |
| cina0 | 0.755 | 0.755 | 0.777 | 0.809 | 0.929 | 0.873 | 0.931 | 0.933 | 0.921 | 0.933 | 0.939 | 0.939 | 0.931 | 0.934 | 0.934 |
| isolet | 0.906 | 0.899 | 0.336 | 0.944 | 0.945 | 0.766 | 0.949 | 0.950 | 0.817 | 0.936 | 0.940 | 0.912 | 0.943 | 0.951 | 0.943 |
| madelon | 0.558 | 0.689 | 0.745 | 0.639 | 0.858 | 0.861 | 0.673 | 0.845 | 0.845 | 0.620 | 0.700 | 0.754 | 0.608 | 0.690 | 0.815 |
| marti0 | 0.881 | 0.881 | 0.881 | 0.874 | 0.874 | 0.874 | 0.870 | 0.870 | 0.870 | 0.878 | 0.870 | 0.866 | 0.879 | 0.868 | 0.854 |
| reged0 | 0.880 | 0.966 | 0.939 | 0.885 | 0.974 | 0.974 | 0.898 | 0.974 | 0.974 | 0.882 | 0.963 | 0.960 | 0.881 | 0.948 | 0.978 |
| secom | 0.935 | 0.935 | 0.930 | 0.935 | 0.931 | 0.931 | 0.934 | 0.932 | 0.932 | 0.935 | 0.933 | 0.929 | 0.935 | 0.930 | 0.928 |
| mnist | 0.564 | 0.823 | 0.525 | 0.959 | 0.966 | 0.905 | 0.968 | 0.970 | 0.938 | 0.964 | 0.966 | 0.953 | 0.966 | 0.971 | 0.968 |
| mnist3v8 | 0.910 | 0.941 | 0.828 | 0.980 | 0.986 | 0.958 | 0.987 | 0.989 | 0.975 | 0.980 | 0.985 | 0.978 | 0.981 | 0.988 | 0.987 |
| mnist4v9 | 0.889 | 0.957 | 0.848 | 0.981 | 0.986 | 0.960 | 0.986 | 0.988 | 0.974 | 0.983 | 0.984 | 0.974* | 0.985 | 0.987 | 0.984* |
| sido0 | 0.970 | 0.972 | 0.953 | 0.973 | 0.968 | 0.968 | 0.974 | 0.969 | 0.969 | 0.972 | 0.973 | 0.973* | 0.973 | 0.974 | 0.960* |
| tis | 0.751 | 0.751 | 0.757 | 0.753 | 0.887 | 0.888 | 0.844 | 0.917 | 0.915 | 0.854 | 0.916 | 0.913* | 0.856 | 0.906 | 0.914* |

Table 2: Average accuracy scores for all methods with specified parameters on original datasets. SRS and RS were computed with 10000 iterations and RF/ET with 10000 trees.

|  | RS | SRS | | ET | |
| --- | --- | --- | --- | --- | --- |
|  | $q = 0.1$ | $q = 0.1$ $\alpha = 0.5$ | $q = 0.1$ $\alpha = 1.0$ | $k = 0.1$ | $k = 1.0$ |
| $q = 0.1$ |  |  |  |  |  |
| RS | — | 1/5/7 | **6/4/3** | 7/2/4 | **6/3/4** |
| SRS$_{\alpha=0.5}$ | **7/5/1** | — | 5/8/0 | 9/2/2 | **10/2/1** |
| SRS$_{\alpha=1.0}$ | 3/4/6 | 0/8/5 | — | 5/2/6 | **6/2/5** |

(a) $q = 0.1 \times p$

|  | RS | SRS | | 
| --- | --- | --- | --- |
|  | $q = 0.01$ | $q = 0.01$ $\alpha = 0.5$ | $q = 0.01$ $\alpha = 1.0$ |
| $q = 0.01$ |  |  |  |
| RS | — | 2/2/9 | 5/2/6 |
| SRS$_{\alpha=0.5}$ | **9/2/2** | — | **7/3/3** |
| SRS$_{\alpha=1.0}$ | **6/2/5** | 3/3/7 | — |

(b) $q = 0.01 \times p$

Table 3: Pairwise t-test (with a significance level of 0.05) comparisons : each element on line $i$ and column $j$ of the table in terms of Win/Draw/Loss is the result of the comparison for method $i$ vs. method $j$: the tree values indicate respectively on how many datasets method $i$ is significantly better / not significantly different / significantly worse than method $j$. All methods were computed with 10000 iterations or trees on all 14 datasets (from Table 1) with parameters specified on columns. In **bold** when the first value is greater than other values.



\end{document}